\documentclass{article}
\usepackage{fullpage}
\usepackage{amsmath,amsfonts}

\usepackage[utf8]{inputenc} 
\usepackage[T1]{fontenc}    
\usepackage{hyperref}       
\usepackage{url}            
\usepackage{booktabs}       
\usepackage{amsfonts}       
\usepackage{nicefrac}       
\usepackage{microtype}      
\usepackage{lipsum}
\usepackage{graphicx}
\graphicspath{ {./images/} }

\usepackage{subcaption}
\usepackage{algorithm}
\usepackage{array}
\usepackage{textcomp}
\usepackage{stfloats}
\usepackage{url}
\usepackage{verbatim}
\usepackage{cite}
\usepackage{algpseudocode}
\usepackage{xcolor}
\usepackage[normalem]{ulem}
\usepackage{changepage}
\usepackage[T1]{fontenc}
\usepackage{lmodern}

\begin{document}
\title{Search-contempt: a hybrid MCTS algorithm for training AlphaZero-like engines with better computational efficiency}

\author{Ameya Joshi\thanks{Independent Researcher, ameya.joshi.s@gmail.com}}


\date{}

\maketitle

\begin{abstract}   
AlphaZero in 2017 was able to master chess and other games without human knowledge by playing millions of games against itself (self-play), with a computation budget running in the tens of millions of dollars. It used a variant of the Monte Carlo Tree Search (MCTS) algorithm, known as PUCT. This paper introduces \textit{search-contempt}, a novel hybrid variant of the MCTS algorithm that fundamentally alters the distribution of positions generated in self-play, preferring more ``challenging'' positions. In addition, search-contempt has been shown to give a big boost in strength for engines in Odds Chess (where one side receives an unfavorable position from the start). More significantly, it opens up the possibility of training a self-play based engine, in a much more computationally efficient manner with the number of training games running into hundreds of thousands, costing tens of thousands of dollars (instead of tens of millions of training games costing millions of dollars required by AlphaZero). This means that it may finally be possible to train such a program from zero on a standard consumer GPU even with a very limited compute, cost, or time budget.

\end{abstract}

\section{Introduction}
\label{sec:introduction}

Conquering chess has been a holy grail and testbed for AI development since its inception. Supercomputer Deep Blue was the first AI system to beat a world champion in chess in classical time control. With development in both hardware and software over the next couple of decades, it became possible to match its level on consumer hardware. However, up until that point human knowledge had to be hard coded into its software to help it play at a high level, an approach adopted by Stockfish, which was the strongest engine at the time. AlphaZero \cite{silver2018general} changed that in 2017 by mastering chess without human knowledge using a very simple search algorithm coupled with reinforcement learning and self-play. It used the same search algorithm as its precursor AlphaGo Zero \cite{silver2017mastering}, i.e., PUCT-based \cite{rosin2011multi} Monte Carlo Tree Search (MCTS). It became the strongest chess engine by playing tens of millions of games against itself and iteratively training a neural network on the policy and value functions resulting from those games. However, generation of the training games requires an immense amount of computation, which costs millions of dollars. 



\subsection*{Contributions}
\label{subsec:contributions}

This paper introduces \textit{search-contempt}, a novel hybrid variant of MCTS that is computationally efficient. The proposed algorithm is an improvement over and a generalization of the PUCT-based MCTS algorithm. It can be used in any self-play based Reinforcement Learning systems such as AlphaGo Zero \cite{silver2017mastering} for go and AlphaZero \cite{silver2018general} for other 2-player games like chess and shogi. 


A comparison between the two algorithms in terms of the variety and quality of training games generated in self-play is also presented. The comparison is performed using a couple of metrics: 
\begin{itemize}
	\item The win-draw-loss distribution, which is a concept that has been widely used with great success by {Leela Chess Zero} \cite{The_LCZero_Authors_LeelaChessZero} for training its neural networks. 

    \item The fraction of repeated games observed in the training games generated. 
\end{itemize}



In regular (non-odds) chess, self-play games generated with {search-contempt} are up to 70 Elo points stronger under conditions suitable for the generation of training games. The proposed search-contempt produces a higher quality of games than PUCT-based MCTS, which means that there is more information content in the self-play games generated. This results in faster progress in playing strength per training game as the neural network is trained. Consequently, a similar level of improvement can be achieved in a fewer number of games. This is the primary reason that search-contempt is more computationally efficient than the PUCT-based MCTS algorithm.

In addition, the similarity of search-contempt to adversarial training and the impact it produces on the distribution of positions encountered during self-play is presented, which could be helpful in making self-play training systems using search-contempt more robust and less prone to adversarial attacks \cite{wang2023adversarial}.

Search-contempt is also shown to gain significant strength (around 150 Elo rating points) over the PUCT-based MCTS algorithm in Odds Chess, a variant of Chess where one side begins the game with a clear disadvantage (i.e., a missing Knight, Rook, or Queen) while the other side plays with all their pieces.




\section{Related Work}
\label{sec:related_work}
{AlphaGo Zero} \cite{silver2017mastering} introduced the concept of lookahead search inside the Reinforcement Learning Training Loop. It performs an MCTS search on each position encountered during self-play. Each MCTS search begins with current position as the root node and grows the search tree by one node with every visit. The path taken along the search tree to add a node is decided by a variant of the PUCT algorithm \cite{rosin2011multi}.
The specific equations, taken from the AlphaGo Zero paper, for the path to take along the tree are: 
\begin{equation}\label{formula:ag0_1_2}
\begin{array}{ll}
     U(s,a) = c_{puct} P(s,a) \frac {\sqrt{\sum_{b} N(s,b)}} {1 + N(s,a)}  \\
     $\quad $ \\
    Q(s,a) = \frac {1} {N(s,a)} \sum_{s'|s,a->s'} V(s') \\
\end{array}
\end{equation}
$\quad $

$\quad $
\begin{equation}
	\label{formula:ag0_3}
a= \underset{b}{\mathrm{argmax}} (Q(s,b) + U(s,b)),
\end{equation}
where $ N(s,a) $ denotes the total number of visits along a certain node $ s $ of the tree after taking action $ a $, $ Q(s,a) $ denotes the average value of the leaf nodes that were reached that involved taking action $ a $ from state $ s $ at some point in its path from the root node, $ c_{puct} $ is a constant, $ V(s) $ is the value function of state $ s $, $ P(s,a) $ is the prior probability of taking an action $ a $ from state $ s $. Here, $P(s,a)$ and $V(s)$ are generated by a neural network trained on self-play games. 
Once the required total number of visits (denoted by $ N $) are complete, a policy $ \pi(a|s_0) $ determines the move to play. This is given by
\begin{equation}
	\label{formula:ag0_4}
\pi(a|s_0)=\frac {N(s_0,a)^{1/\tau}} {\sum_b N(s_0,b)^{1/\tau}}.
\end{equation}
The variable $ \tau $, which represents the temperature of a move, does not affect the MCTS search itself but considerably affects the trajectory of the positions encountered during a game of self-play. To measure the current strength of the engine, AlphaGo Zero used a low value of $ \tau $, i.e., $ \tau_{1} $ that results in the best move always being played. This produces very high quality games but are limited in variety as is expected from strong engines. However, this is not ideal as training data. To generate variety in training games, the authors choose a high value of $ \tau $, i.e., $ \tau_2 $, which selects with high probability even moves that may not be the best but are good for exploration. This, however, weakens the level of play in training games.


With great success, the same PUCT-based MCTS algorithm has been subsequently used for {AlphaZero} as well 
\cite{silver2018general}. However, the problem of weak level of play for the training games in order to generate suffcient variety persists. Search-contempt, which is a generalization of the PUCT-based MCTS algorithms used in the past, aims to mitigate this problem. 

\section{Proposed {search-contempt} Hybrid MCTS Algorithm}
\label{sec:proposed_algo}

The proposed hybrid asymmetric MCTS search algorithm, i.e. \textit{search-contempt} is a mixture of two components. The first component is PUCT as used in the AlphaZero paper. The second component is Thompson Sampling(TS). Search-contempt is characterized by a new integer parameter $ N_{scl} $ (a short hand for search-contempt-node-limit), which defines the search in the following way. For the root node and all nodes corresponding to the player-to-move, i.e., nodes with even depth ($ d_{s} = 0,2,4,...$), the PUCT algorithm is used to select child nodes to explore. For all the nodes corresponding to the opponent, i.e., nodes with odd depth ($ d_{s} = 1,3,5,...$) the PUCT algorithm applies until the total nodes visited in the subtree of that node reaches $ N_{scl} $. At this point, the visit distribution of its child nodes is frozen and stored. The child nodes to visit for any subsequent visits beyond $ N_{scl} $ is sampled proportionally to the frozen visit distribution which is TS, the second component. In other words, after $ N_{scl} $ visits, the visit distribution remains fixed, irrespective of how good or bad the child nodes turn out to be. This is in contrast to the PUCT algorithm which adjusts the subsequent visit distribution based on how good or bad the child nodes turn out, with more visits assigned to nodes that turn out to be ``good'' and less visits assigned to nodes that turn out to be ``bad''.



A mathematical description of search-contempt follows. All notation and parameters retain the same meaning as described in the AlphaGo Zero paper. Only four
new notations are used: $ d_{s} $, $ N_{scl} $, $ N_{scl}(s,a) $, and $ P_{mcts}(s,a) $. Here, $ d_{s} $ is the depth of the node to explore. $N_{scl} $ is a new integer search parameter, which is the visit count at which the search transitions from PUCT to Thompson Sampling. $ N_{scl}(s,a) $ is the ``frozen'' node count associated with a particular action (i.e., a child node). $P_{mcts}(s,a) $ is the probability of selecting an action $a$ from state $s$ during a single visit of MCTS. Note that $ P_{mcts}(s,a) $ is different from both $ P(s,a) $ (which is the prior probability of selecting an action) and $ \pi_{a} $ (which is the probability of selecting an action at the root node for actually playing a move once the search is complete).

\begin{figure}[t]
\center
  \includegraphics[scale=0.42]{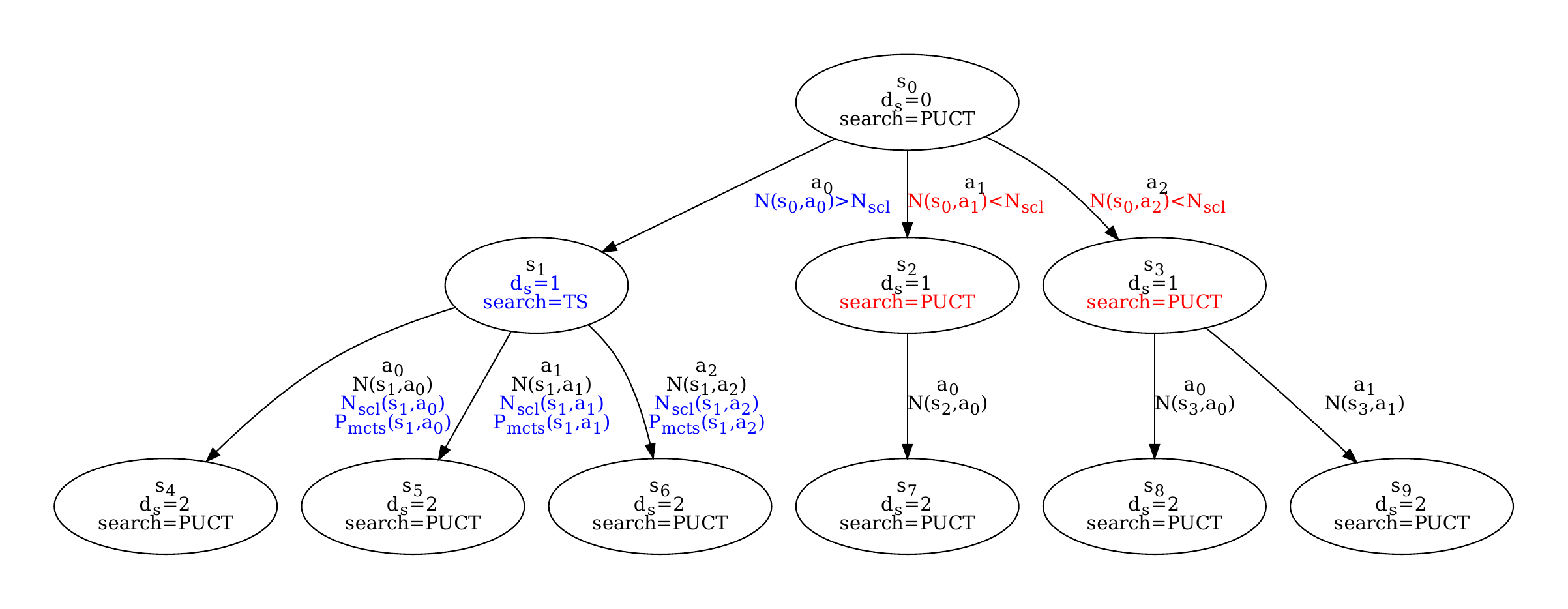}
  \caption{This is a sample snapshot of the state of the tree search using search-contempt. State $ s_0 $ is the root node. State $s_{1}$ has transitioned to Thompson Sampling (TS) since its $ d_s $ value is odd and its visit count $ N(s_{0},a_{0}) $ exceeded $ N_{scl} $ at some point during the search. For $s_{1}$ and all other states which have transitioned their search to TS, the edges of the node separately store $ N_{scl}(s,a) $ and $ P_{mcts}(s,a) $ for each of its actions. These values were frozen at the point where $ N(s_0,a_0) = N_{scl} $ which translates to $ (N_{scl}(s_1,a_0) + N_{scl}(s_1,a_1) + N_{scl}(s_1,a_2) == N_{scl})$ always being valid. States $ s_2 $ and $ s_3 $ on the other hand still have their search as PUCT since their visit count has not exceeded $ N_{scl} $ yet. State $ s_{0} $'s visit count has exceeded $ N_{scl} $ but it still uses PUCT since its $ d_s $ is even, i.e., it is a node corresponding to the player whose turn it currently is. Similar reasoning applies to states $ s_4 $ - $ s_9 $ which, therefore, also use PUCT.}
  \label{fig:mcts_tree}
\end{figure}

$ N_{scl}(s,a) $ is defined in terms of $ N_{scl} $ as
\begin{equation}
	\label{formula:hyb_1}
N_{scl}(s,a) = N(s,a) \text{ when } \sum_{b} N(s,b) = N_{scl},
\end{equation}
where $ N_{scl} $ is the new search parameter that defines the visit count at which the search transitions from PUCT to Thompson Sampling. $ N_{scl} $ is a constant for all nodes during search. $ N_{scl}(s,a) $ is the ``frozen'' visit count associated with a particular action $ a $ and state $ s $. Using the above definition of $ N_{scl}(s,a) $, $ P_{mcts}(s,a) $ can now be expressed as:
\begin{itemize}
    \item If $(d_{s}mod2 = 0)$  or $ (\sum_{b} N(s,b) \leq  N_{scl})$, then
\begin{equation}
	\label{formula:hyb_2}
P_{mcts}(s,a)=
    \begin{cases}
        1 & \text{if } a= \underset{b}{\mathrm{argmax}} (Q(s,b) + U(s,b))\\
        0 & \text{if } a \neq \underset{b}{\mathrm{argmax}} (Q(s,b) + U(s,b)),
    \end{cases}
\end{equation}

\item If $ (d_{s}mod2 = 1) $ and $(\sum_{b} N(s,b) > N_{scl}) $, then 
\begin{equation}
	\label{formula:hyb_3}
P_{mcts}(s,a) = \frac {N_{scl}(s,a)} {\sum_{b} N_{scl}(s,b)} = \frac {N_{scl}(s,a)} {N_{scl}}.
\end{equation}
\end{itemize}
Here, $ P_{mcts}(s,a) $ is the probability of selecting an action $ a $ from state $ s $ during a single visit of MCTS, and $ d_{s} $ is the depth of the state $ s $ from the root of the tree. Equation \ref{formula:hyb_2} is the PUCT component, while Equation \ref{formula:hyb_3} is the Thompson Sampling component of search-contempt.

All other aspects of the algorithm are identical to the AlphaGo Zero paper \cite{silver2017mastering}. Note that when $ N_{scl} $ is set to $ \infty $, Equations \ref{formula:hyb_2} and \ref{formula:hyb_3} reduce to Equation \ref{formula:ag0_3}, which is equivalent to the PUCT-based MCTS search used in AlphaGo Zero.

Figure \ref{fig:mcts_tree} shows a snapshot of a possible search tree state with search-contempt and demonstrates the application of Equations \ref{formula:ag0_1_2} to \ref{formula:hyb_3} indicating the conditions under which a node transitions from PUCT to Thompson Sampling. Search-contempt markedly affects the visit distribution during the search on each move, which in turn affects the type of positions visited and the moves played during training game generation, as detailed in the following sections.

\paragraph{Implementation.}
The code for search-contempt is available at \url{https://github.com/amjshl/lc0\_v31\_sc}, which also contains the instructions to reproduce the results of the self-play and match games described in the next section. This branch is committed from a clone of version 31 \cite{lc0_v31} of the Leela Chess Zero project, an open source implementation \cite{The_LCZero_Authors_LeelaChessZero} of the AlphaGo Zero family of engines.

\section{Experimental Results for {search-contempt}}
\label{sec:exp_results}
In this section, we present the experimental results that highlight the benefits of the proposed search-contempt over the PUCT-based MCTS, not only in terms of improved playing strength of a fully-trained system, but also the increased quality of the self-play games generated, used for efficiently training the reinforcement learning based systems.

This section begins with the comparison of performance between the proposed search-contempt and the PUCT-based MCTS algorithm in the context of Odds Chess. Subsequently, we show how the new parameter $ N_{scl} $ in search-contempt can be used to control the win-draw-loss distribution of self-play games which are used for training a neural network in the context of reinforcement learning in regular chess. The role of temperature $ \tau $ in the PUCT-based MCTS approaches to control the win-draw-loss distribution is also experimentally shown. The ideal conditions for generating self-play training games with both the approaches are compared in terms of their relative playing strength, which is a measure of the quality of training games produced. The impact of search-contempt on the distribution of positions encountered in training games is discussed qualitatively with the help of a sample chess position encountered during self-play. Finally, a plausible training schedule for training a neural network using search-contempt with self-play is presented, which is expected to use significantly less computation power compared to the PUCT-based MCTS search.

We begin with a comparison in performance between the proposed search-contempt and the PUCT-based MCTS search in the context of Odds Chess.

\subsection{Rating Improvement in Odds Chess}
\label{subsec:odds_chess}

Odds chess is a variant of chess where one of the players, the odds giver, has a severe handicap of a missing piece (a missing Knight, Rook, or Queen) while the other player, the odds receiver, plays with the full set of pieces. As expected, the odds giver is the stronger player of the two, so that the match is even. Before 2023, even the top engines were not able to consistently beat high-rated players (> FIDE 2000) in odds chess. This changed recently with several innovations, discussed in this blog \cite{lc0_blog} by the Leela Chess Zero team, like wdl contempt \cite{wdl_contempt}, and specially trained Knight Odds \cite{lc0_lko} and Queen Odds \cite{lc0_lqo} network, causing a large increase in playing strength and more notably a change in playing style, making Leela Chess Zero the strongest rated engine for odds chess in history.

However, the previous versions of the Leela Odds engine use PUCT-based MCTS as its search algorithm, which suffers from a strange phenomenon. The odds play gets better as more nodes are searched, but only up to a certain point, after which the search outcome or playstyle begins to resemble the strongest engines in non-odds chess which are weaker in odds chess. Hence, the total nodes searched per move is limited in order to optimize playing strength for odds chess. This limit is around 800 nodes for queen odds, 10\,000 nodes for rook odds, and 20\,000 nodes for knight odds chess.

\begin{table}[t]
  \begin{center}
    \caption{Comparison between search-contempt and PUCT-based MCTS for performance in Queen Odds Chess against a fixed opponent.}
    \label{tab:1}
    \begin{tabular}{c|c|c} 
    \toprule	    
	    \- & Win Percentage(\%) & Win Percentage(\%) \\
	    Nodes Searched & PUCT-based MCTS & search-contempt \\
      \midrule
	    100 & 23.43 & \textcolor{blue}{36.76}\\
	    800 & 35.25 & \textcolor{blue}{57.81}\\
	    20\,000 & \textcolor{red}{25.40} & \textcolor{blue}{81.73}\\
    \bottomrule	    
    \end{tabular}
  \end{center}
\end{table}

The performance of the PUCT-based MCTS and search-contempt for various total nodes searched in Queen Odds chess is presented in Table \ref{tab:1}. For PUCT-based MCTS, as the nodes searched at the root is increased, win percentage against a fixed opponent first increases, reaches a maximum of 35.25\% at 800 nodes and then drops to 25.40\% on further increase. In contrast, the win percentage of search-contempt is monotonically increasing, and does not suffer from this effect. Also, for the same node count, search-contempt outperforms PUCT-based MCTS by a significant margin. For instance, at 800 nodes per move, using similar computation, search-contempt achieves a win percentage of 57.81 as opposed to 35.25 for PUCT-based MCTS. This indicates that search-contempt exhibits not only better absolute performance but also better scalability with the number of nodes searched.


The reason that search-contempt does not suffer from this effect is that by freezing the distribution of visits after $ N_{scl} $ visits for the opponent's(player 2) turn, we maintain the imperfect play expected of the weaker opponent while simultaneously allowing for higher number of visits for the side giving odds (i.e., player 1), thereby increasing the strength of player 1 relative to player 2. With the inclusion of search-contempt, the leela family of odds engine bots gained an additional 150 Elo rating against humans online. These bots can be challenged on lichess \cite{lichess_lqo}.


\subsection{Compute-Efficient AlphaZero with search-contempt}
\label{subsec:sc_results}

AlphaGo Zero trained on upto 29 million games for their larger model, while AlphaZero was trained on 44 million games to exceed the level of Stockfish 8, the reigning TCEC chess champion at the time. The estimated cost of training such an engine was likely in the range of tens of millions of dollars. There is a possibility that this can be significantly reduced by using search-contempt instead of the PUCT-based MCTS as used by AlphaZero or the open source Leela Chess Zero. We show this by comparing the way in which the PUCT-based MCTS and search-contempt can adjust its parameters to tune the win-draw-loss (w-d-l) distribution of the training games it could possibly generate during self-play and also showing how for a particular distribution of w-d-l, search-contempt results in stronger play compared to PUCT-based MCTS. For the game of chess, which is almost symmetrical with respect to both the players and which is dominated by draws at the top level, we find that the quantity $ \frac{w+l} {d} $ effectively captures the w-d-l distribution into a single quantity. For games where draws are absent like go, this does not work and some other appropriate quantity that better captures the variety in positions visited or the game outcome would have to be defined.

\begin{figure}[t]
\centering
  \includegraphics[scale=0.6]{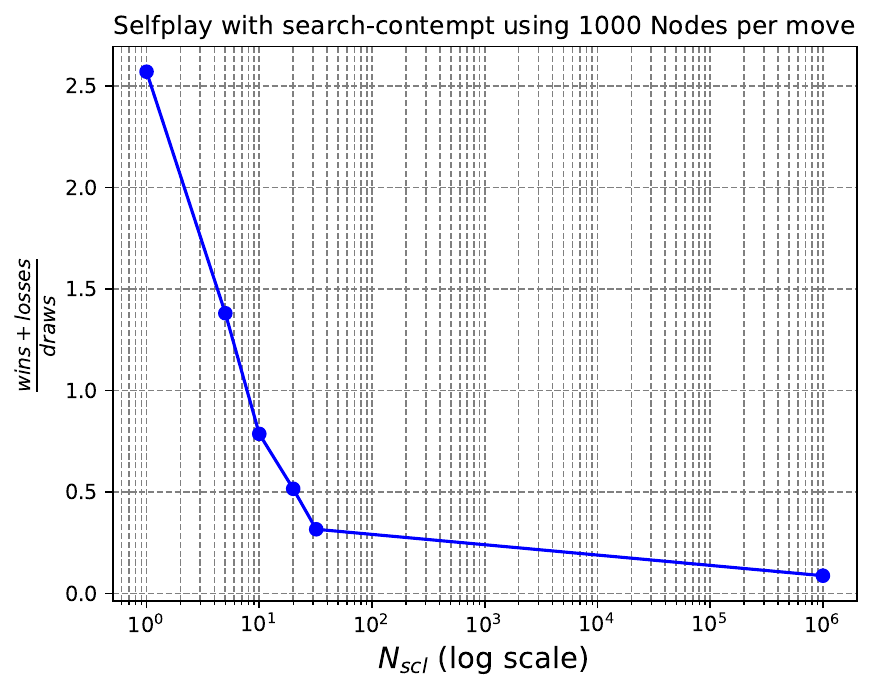}
	\caption{This shows the plot of $ \frac {w+l} {d} $ vs $ N_{scl} $ for the search-contempt algorithm in self-play mode. Here, $ w $, $ l $ and $ d $ is the number of wins, losses and draws respectively from the white player's perspective. The total node counts used is 1000 for each move and a total of 100 games are played out for each value of $ N_{scl} $. There is a value of $ N_{scl} $ which is optimal for generating training games, which occurs roughly at $ N_{scl} = 5 $ when $ \frac {w+l} {d} \approx 1 $.} 
  \label{fig:wdl_vs_scl}
\end{figure}

\subsubsection{Impact of $ N_{scl} $ on win-draw-loss distribution in training games with search-contempt}
\label{subsubsec:nscl_vs_wdl_subsubsec}

 $ N_{scl} $, the new parameter introduced in the {search-contempt} algorithm, can act as an additional lever to vary, both the distribution of positions visited in the training games that are generated and also the distribution of the outcome of the games, i.e. the percentage of wins ($w$), draws ($d$) and losses ($l$). This is in addition to the $ \tau $ parameter used in AlphaGo Zero. 

Figure \ref{fig:wdl_vs_scl} shows how the $w$-$d$-$l$ distribution is affected by the choice of $ N_{scl} $ for chess. Note that all values of $ N_{scl} $ use the same amount of total computation as that of the PUCT-based algorithm. For $ N_{scl} = 1\,000\,000 $, which is equivalent to the PUCT-based MCTS, the quality of play is high, but $ \frac {w+l} {d} $ is low as most games are drawn since both sides assume optimal search for themselves as well as their opponents. Since training games require a roughly equal distribution of $w$-$d$-$l$, this is not the setting that can used for generating self-play games. However, as $ N_{scl} $ value gets less than 50, the fraction of wins and losses starts increasing, reaches the value of 1 at $ N_{scl} = 8 $ and goes above 2.5 when $ N_{scl} = 1 $. The ideal ratio occurs at $ N_{scl} = 5 $, where $ \frac {w+l} {d} \approx 1 $, which is well suited for training. The reason for this variation in $ \frac {w+l} {d} $ is that with decreasing $ N_{scl} $, both players assume their opponent gets weaker and weaker, and therefore, starts placing importance to suboptimal moves, which are good against weak opponents but could be objectively bad. This causes weaker play, and thus, more wins and losses compared to draws. This is a tradeoff that is necessary for generating enough exploration during training game generation. Note that with asymmetric MCTS variants like search-contempt, although both the players (for self-play) use the same value of $ N_{scl} $ in a particular game, they cannot share their search tree, unlike in the case of PUCT-based MCTS. Therefore the search tree has to be cleared after every move. The self-play games that were used in Figure \ref{fig:wdl_vs_scl} are linked below in the pgn format.
\begin{itemize}
    \item $ N_{scl} = 5$: \url{https://github.com/amjshl/compute_efficient_alphazero/blob/master/files/hybrid_selfplay_Nscl_5_N_1000.pgn}.
\item $ N_{scl} = 10$: \url{https://github.com/amjshl/compute_efficient_alphazero/blob/master/files/hybrid_selfplay_Nscl_10_N_1000.pgn}.
\item $ N_{scl} = 32$: \url{https://github.com/amjshl/compute_efficient_alphazero/blob/master/files/hybrid_selfplay_Nscl_32_N_1000.pgn}. 
\item $ N_{scl} = 1\, 000\,000$: \url{https://github.com/amjshl/compute_efficient_alphazero/blob/master/files/hybrid_selfplay_Nscl_1000000_N_1000.pgn}.

\end{itemize}

Qualitatively, the games are of a completely different nature with different $ N_{scl} $ values. For high values of $ N_{scl} $, the games are of high quality but very limited in variety. Most games are similar and both sides are quick to exchange off pieces and liquidate to a drawn endgame. This resembles the quality of games that is expected of top engines performing search at $1000$ nodes per move. As $ N_{scl} $ gets closer and closer to the optimal value of 5, the nature of the games completely changes. The quality reduces but almost every game features a lot more complications involving piece imbalances, trapped pieces, perpetual checks, long-term piece sacrifices, dynamic stalements, desperado tactics, checkmates at the middle of the board, and many more 2 or 3-move tactics. Almost every position has something interesting about it.  

Here is an explanation for the shift in the nature of the games as $ N_{scl} $ is reduced. This parameter clearly alters the visit distribution favoring exploration of the moves for which the opponent (which in the case of self-play is itself) misevaluates the position at lower nodes compared to higher nodes. The greater the difference in evaluation, the more likely the move is to be explored. Not only does search-contempt play out unusual positions, it actively seeks out positions which the network is bad at evaluating or searching, which are usually absent or less frequent in the games generated using PUCT-based MCTS. A quick scan of the self-play games suggests that the frequency of ``interesting'' positions is about 20-30 times higher with the lower values of $ N_{scl} $.

\subsubsection{Impact of temperature ($ \tau $) on win-loss-draw distribution in training games}
\label{subsubsec:tau_vs_wdl_subsubsec}

The way AlphaZero using PUCT-based MCTS achieves variety in the positions and outcome of self-play training games is fundamentally different from that achieved with search-contempt. It primarily relies on temperature $ \tau $ as a lever for changing the $w$-$d$-$l$ distribution of training games. Here, $ \tau $ is the temperature parameter from the AlphaGo Zero paper which determines the policy $ \pi_{a} $ from $ N(s,a) $, after the search is completed for each turn. The formula used is $ \pi_{a} \propto N(s,a)^{1/\tau} $. 

\begin{figure}[h]
\center
  \includegraphics[scale=0.6]{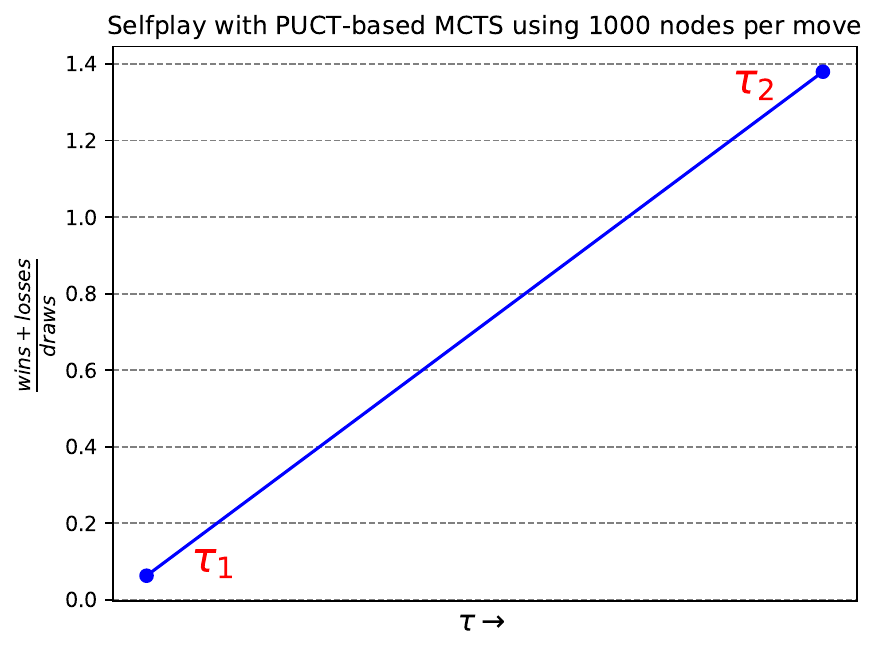}
	\caption{This shows the plot of $ \frac {w+l} {d} $ vs $ \tau $ for self-play with the PUCT-based MCTS algorithm using 1\,000 visits for each turn. 100 games are played for each value of $ \tau $. The exact values of $ \tau $ used is omitted since this value is not fixed for all the moves in a game and so is just a one-dimentional projection of the $ \tau $ schedule used for each case. All that can be said is that $ \tau_{2} > \tau_{1} $ for any particular move number. Here, $ \tau_2 $ gives a $ \frac {w+l} {d} $ ratio of 1.38 compared to just 0.05 for $ \tau_1 $. $ \tau_2 $ is thus a better candidate for generating training games than $ \tau_1 $.} 

  \label{fig:wdl_vs_temperature}
\end{figure}

The experiments using different values of $ \tau $, i.e., $\tau_1$ and $\tau_2$ have been recreated and presented in Figure \ref{fig:wdl_vs_temperature}. The lower temperature $ \tau_1 $ produces high-quality games but results in mostly draws and low values of $ \frac  {w+l}{d} $, while the higher temperature $ \tau_2 $ trades off the quality of play for more variety and a better $w$-$d$-$l$ distribution ($  \frac  {w+l}{d} \approx 1 $). Not surprisingly, this prompts the use of $ \tau_1 $ for optimizing strength in competitive matches and the use of $ \tau_2 $ for generation of training games.

The complete set of the games using PUCT-based MCTS with different $ \tau's $ in Figure \ref{fig:wdl_vs_temperature} are linked below:
\begin{itemize}
    \item $\tau_1$: \url{https://github.com/amjshl/compute\_efficient\_alphazero/blob/master/files/pure\_PUCT\_tau\_1.pgn}. 
\item $\tau_2$: \url{https://github.com/amjshl/compute\_efficient\_alphazero/blob/master/files/pure\_PUCT\_tau\_2.pgn}.
\end{itemize}

\subsubsection{Varying $ N_{scl} $ results in better quality of training games than varying $ \tau $ }
\label{subsubsec:tau_vs_nscl_match}

It turns out that varying $ N_{scl} $ is a much more effective way of generating a balanced $w$-$d$-$l$ distribution than is $ \tau $, while generating training games with self-play. These two self-play methods are compared with configurations that are individually tuned to be suitable for training games. For a fair comparison, the $ \frac {w+l} {d} $ ratio close to 1, ie 1.38 is chosen for both the cases. Search-contempt, using $ N_{scl} =5 $ and $ \tau_1 $ outperformed PUCT-based MCTS, using $ \tau_2 $ by 70 rating points despite using similar computational power. The details are given in Table \ref{tab:tab3}. This means that the quality of training games generated by tuning $ N_{scl} $ is greater than that generated by just tuning $ \tau $, which in turn means that networks can improve at a faster rate during training and achieve a higher ceiling. Also as training progresses, the value of  $ N_{scl} $ at which the w-d-l ratio is ideal, is expected to increase and therefore the rating difference of self-play games is expected to increase beyond even 70 points. The games from Table \ref{tab:tab3} in pgn format is linked at \url{https://github.com/amjshl/compute\_efficient\_alphazero/blob/master/files/pure_PUCT_tau_2_vs_hybrid_Nscl_5.pgn}.

\begin{table}[t]
  \begin{center}
	  \caption{ Match between (a) PUCT-based MCTS using $ \tau_2 $ and (b) search-contempt using $ \tau_1 $ 
      and $ N_{scl} = 5 $ with both players using 1\,000 visits for each turn. The values of $ \tau $ and $ N_{scl} $ are 
      chosen such that $ \frac{w+l} {d} $ ratio of 1.38 is produced for each case in self-play mode. There is a clear improvement of 70 Elo rating points for (b) over (a).}

    \label{tab:tab3}
    \begin{tabular}{c|c|c|c|c|c|c} 
    \toprule	    
	    Temperature schedule & Search-type & $ N_{scl} $ & Wins & Draws & Losses & Rating \\
\midrule
        $ \tau_2 $ & PUCT-based MCTS & - & 17 & 46 & 37 & 0 \\
	    $ \tau_1 $ & search-contempt & 5 & 37 & 46 & 17 & 70.43 \\
    \bottomrule	    
    \end{tabular}
  \end{center}
\end{table}


\begin{figure}[H]
\center
  \includegraphics[scale=0.6]{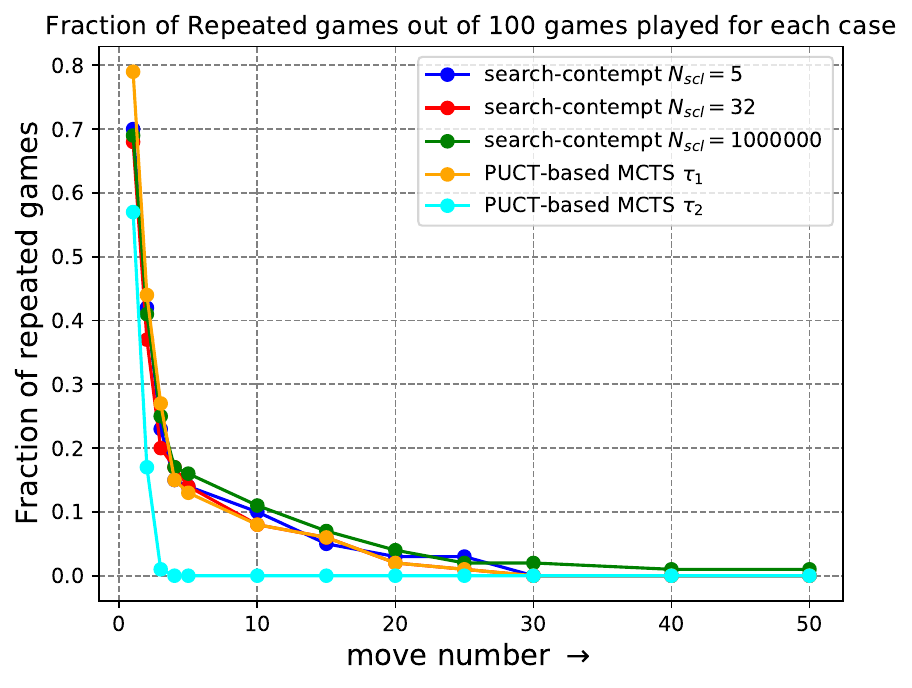}
	\caption{This is a plot of the fraction of repeated games out of a total of 100 games vs move number for each of the self-play experiments presented earlier. A value of 1 means all the 100 games are identical upto that particular move number and a value of 0 means none of the 100 self-play games are repeated. For the case of high temperature, ie $ \tau_2 $, the repeat rate drops very quickly to 0 by move 4. For $ \tau_1 $ on the other hand, the repeat rate drops more slowly and all the games are unique by move 30 which is expected since $ \tau_2 > \tau_1 $. For search-contempt, which uses $ \tau_1 $ for self-play, the repeat rate is not very different from the PUCT-based MCTS case with $ \tau_1 $, even for different values of $ N_{scl} $, demonstrating that lowering $ N_{scl} $ does not negatively impact the repeat rate. Note that for all the cases the games are almost all unique by move 20.} 

  \label{fig:repetition_vs_moves}
\end{figure}

Despite the better quality of play with search-contempt, $ N_{scl} $ cannot completely replace the role of temperature $ \tau $ . Temperature would still be required in training in order to avoid repetition of games. However, the ability to control $ \frac {w+l} {d} $  using $ N_{scl} $ in addition to $ \tau $ enables much more flexibility in the choice of temperature schedule in training. The temperature lever can be adjusted to get a good variety early on in the game while still maintaining a high draw rate and thus the quality of games. Once $ \tau $ is determined to give sufficient variety, $ N_{scl} $ could then be used to tune the $ \frac{w+l} {d} $ ratio to close to 1, since its negative impact on the level of play is lower than that of $ \tau $. The fact that the value of $ N_{scl} $ does not increase the number of repetitions in the training games generated during self-play is illustrated in Figure \ref{fig:repetition_vs_moves}.

\subsubsection{Qualitative impact of $ N_{scl} $ on the distribution of training games }
\label{subsubsec:nscl_impact_training_games}

\begin{figure}
  \includegraphics[width=\linewidth]{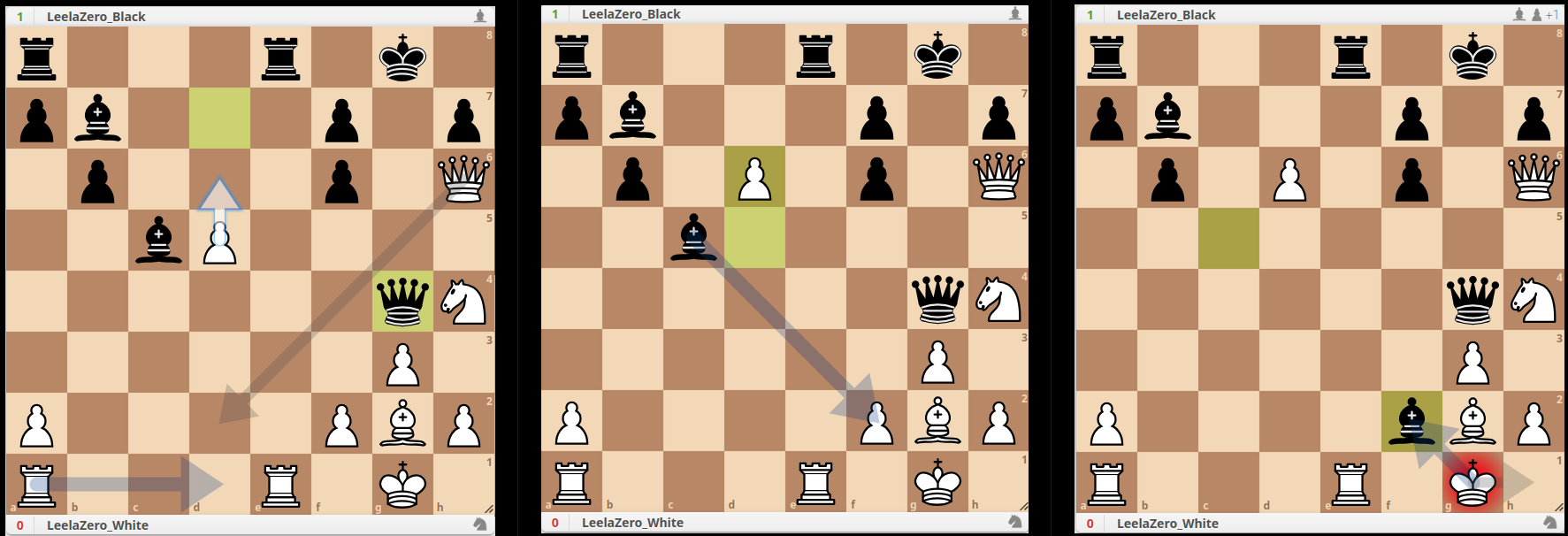}
	\caption{ This shows 3 consecutive chess positions from a self-play game (game 1 of $ N_{scl} = 5 $) played using search-contempt and how it deviates from PUCT-based MCTS. Position 1 is roughly balanced with the objective evaluation, Q of 0.0 with the best move of Rd1. (Here Q varies from -1 to 1, with Q=1 referring to white winning, Q=-1 referring to black winning and Q=0 referring to draw or an equal position for both the players). This is what the PUCT-based MCTS prefers as expected. However search-contempt goes for d6 since at the low node count of 5, the evaluation of Position 2 is is slightly favorable for white. This is because the raw neural network (NN) evaluates Position 2 as 0.004, which is a severe misevaluation since it is an objectively losing position with a Q of -0.82. It is only at Position 3 that the NN evaluates it around -0.34 which is closer to the objective value.} 

  \label{fig:chess_position}
\end{figure}

One interesting instance of how search-contempt 
changes the distribution of the training games generated is shown in Figure \ref{fig:chess_position}. It favors the visit of a position that is misevaluated as favorable for white when in fact it is objectively favorable for black. This gets to the heart of the distribution in the training data. Since PUCT-based MCTS does not visit the slightly ``tricky'' Position 2, such positions are pruned from the set of visited positions during self-play. However with search-contempt, such positions with grave NN misevaluations are actively sought out so that they are more likely to be fixed in the next iteration of training. In effect, search-contempt with sufficiently low values of $ N_{scl} $ can be said to perform a self-adversarial attack on its own policy and value misevaluations, favoring the exploration of those kinds of positions over the objectively best moves. These positions are usually what humans would qualitively classify as ``interesting'', ``complicated'', ``dynamic'', or ``unusual''.

The change in distribution of training positions enables search-contempt to function as a more efficient policy and value improvement operator than PUCT-based MCTS.

\subsubsection{ Plausible training schedule with search-contempt }
\label{subsubsec:sc_training_schedule}

The significant impact, that the limited number of training games (about 200\,000) has on the strength and playing style of the specialized Odds Chess Networks (which are fine-tuned versions of Leela Chess Zero Networks for non-odds chess), points to the fact that with careful control on the training schedule and parameters, very quick progress on playing strength can be made if the training data is generated using search-contempt. Likely using similar training schedules and number of games as the Odds Networks, playing strength rivalling that of the current best networks can likely be achieved for standard chess as well. This amount of compute seems well within the reach of modern consumer hardware.

\begin{figure}[t]
\center
  \includegraphics[scale=0.6]{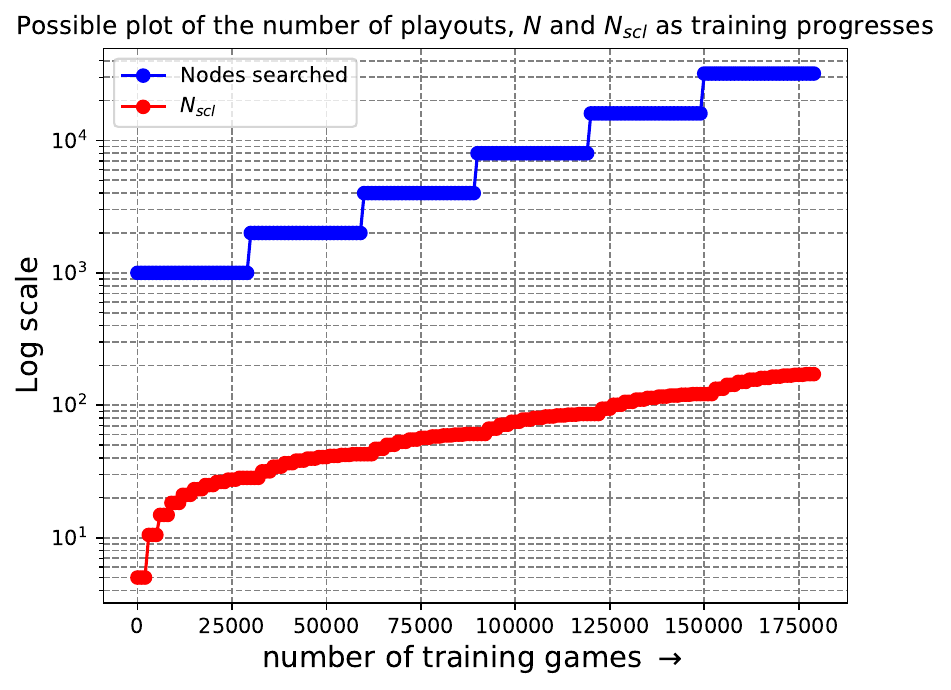}
	\caption{ This is a candidate schedule for training a neural network with the search-contempt algorithm starting with the correct best network. Note the log scale on the y-axis. Since for the latest network used for generating self-play games, $ N_{scl} = 5 $ gives a $ \frac {w+l}{d} $ ratio of roughly 1, training could begin with N=1000 total visits per move and $ N_{scl} = 5 $. After around 3\,000 training games have been generated with this configuration, the network can be trained a certain number of steps which will change the w-d-l distribution. With the resulting network, a new $ N_{scl} $ value, likely higher than the previous one would be required so as to the maintain a consistent $ \frac {w+l}{d} $ ratio. Eventually, $ N_{scl} $ will settle to some constant value. This happens at 30\,000 games in this speculative figure. At his point N can be doubled and $ N_{scl} $ can be found which again maintains the $ \frac {w+l}{d} $ ratio at roughly 1. Repeating this process will result in increasingly stronger networks. } 

  \label{fig:training_schedule}
\end{figure}

 A plausible training schedule using search-contempt to train increasingly stronger networks is indicated in Figure \ref{fig:training_schedule}. With the amount of computation available for large-scale projects such as AlphaZero and Leela Chess Zero, the limits of peak strength of chess engines can be further stretched with search-contempt compared to PUCT-based MCTS.

\section{Conclusion}
\label{sec:conclusion}

 To conclude, search-contempt, which is a hybrid asymmetric generalization of PUCT-based MCTS, previously used in self-play based reinforcement learning framework by AlphaZero, is introduced in this paper. Search-contempt has been shown to give a significant boost in strength to the Leela Chess Zero engine in Odds chess, and is also the search currently used in the leela odds family of bots in lichess.

 In addition, we have shown that search-contempt is a much better way of generating high-quality training data for reinforcement learning by self-play in regular chess as well. The $ N_{scl} $ parameter serves as a more effective lever for adjusting the $w$-$d$-$l$ ratio so that the $ \tau $ used by AlphaGo Zero can instead be used as a lever to add the required amount of variety to the training games. This is demonstrated by a match between two different configurations of search, both of which give a reasonable $w$-$d$-$l$ distribution. The configuration which uses a low value of $ N_{scl} $ in search-contempt instead of a high value of $ \tau $ in the PUCT-based MCTS search, results in stronger play by 70 Elo rating points. 

 Unifying the observed play with search-contempt in both odds chess and regular chess , we can conclude that search-contempt can be used to improve play wherever the opponent has inaccuracies in its play and does not have access to perfect policy and value evaluation functions. In the case of odds chess, we know that humans play imperfect chess, and modern engines in regular chess, although superhuman, play far from perfect chess and regularly make inaccuracies in complicated positions. Thus, it is likely there is still plenty of room for improvement.

 We have seen that search-contempt actively seeks out complicated positions which are evaluated incorrectly at lower values of total nodes searched but which are more accurately evaluated at higher values of total nodes searched. This can be a very efficient and automated way of generating tactical puzzles in chess at various difficulty levels. As the values of $N_{scl}$, the total nodes searched at the root $N$, and the absolute difference between $N_{scl}$ and $N$ increase, the likelihood of search-contempt uncovering chess puzzles with greater levels of difficulty also rises. These are positions which are misevaluated at $ N_{scl} $ nodes searched but correctly evaluated at $ N $ nodes searched.

The type of games generated by search-contempt suggests that the space of even the ``reachable'' chess positions explored currently by the PUCT-based MCTS search, although way better than a few years ago, is likely only a small fraction of that which can possibly be explored with search-contempt. Filling out these ``holes'' in the training data could (a) improve peak playing strength, (b) reduce the computation requirement to a point where existing strength could be achieved with networks trained from zero on consumer hardware at a fraction of the cost, and also (c) prove to be a means of reducing the susceptibility of such self-play trained systems to adversarial attacks \cite{wang2023adversarial}.

\section{Acknowledgements}
\label{sec:acknowledgements}

This work is heavily inspired by the recent progress in odds chess by the Leela Chess Zero (lc0) developer community. Huge credit to Noah, Naphthalin, marcus98, Hissha, and many others from the lc0 discord server for the highly insightful discussions and valuable contributions to odds chess play, which served as inspiration for the idea of search-contempt. The implementation itself would not have been possible without the lc0 codebase made freely available to the public, provided by the Leela Chess Zero team.  Also, thanks to some other users like Rust who ran independent testing of the strength improvement in odds chess and helped point out bugs in the code. Also, thanks to those lc0 developers who have painstakingly written, optimized, and refined the lc0 code to a really high degree of performance which allowed for such fast execution of the experiments that would otherwise have been impossible, given the limited availability of hardware.

\bibliography{myreferences}
\bibliographystyle{unsrt}

\end{document}